\newcommand{\fighere}[1]{
\begin{centering}
*************************************\\
Figure~\ref{#1} goes about here \\
*************************************\\
\end{centering}
}
\newcommand{\tabhere}[1]{
\begin{centering}
*************************************\\
Table~\ref{#1} goes about here \\
*************************************\\
\end{centering}
}

\documentstyle[epsfig,11pt,a4]{article}
\title{Multi-level post-processing for Korean character recognition 
using morphological analysis and linguistic evaluation\thanks{This project
was supported by Korea Science and Engineering Foundation (KOSEF) with grant number
92-21-00-05.}}
\author{Geunbae Lee, Jong-Hyeok Lee, JinHee Yoo \\
Department of Computer Science \& Engineering \\
Pohang University of Science \& Technology \\
San 31, Hoja-Dong, Pohang, 790-784, Korea \\
{\tt gblee@vision.postech.ac.kr, jhlee@vision.postech.ac.kr}}
\date{}

\begin{document}

\maketitle

\clearpage

\begin{center}
{\bf\Large Summary}
\end{center}
Optical character recognition has been actively researched as convenient
means of automatic data input to computers. However, due to excessive
similarities among recognized characters and noises in images, there
have been limitations to direct performance improvements of character
recognition method. So post-processing is always required for practical
character recognition.  Previous post-processing methods use only
within-word contextual information
such as character transition and confusion probabilities.
In contrast, we extend the concept of contextual information to the
sentence level, and present a
multi-level post-processing method that utilizes linguistic information
including character, word,
syntax, and even semantic-based knowledge for domain-independent off-line
text recognition. The proposed post-processing system
performs three-level processing: candidate character-set selection,
candidate {\em eojeol} (Korean word) generation through morphological
analysis, and final single {\em eojeol}-sequence selection by high-level linguistic
evaluation.
The candidate selection restricts the number of candidates for later
processing, and supplements the candidate sets by adding similar characters
for
error correction. The morphological analysis uses word-fragment level constraints to
filter out erroneous recognition results by checking if the recognized
character sequences can form a grammatically correct {\em eojeol}. The
linguistic evaluation uses syntax and semantic-level statistical information
to further filter out erroneous results.
We utilized two high-level linguistic constraints for our linguistic
evaluation: tri-gram part-of-speech tagging and mutual information
based co-occurrence relations.  All the required linguistic
information and probabilities are automatically acquired from a 
statistical corpus analysis.  Experimental results demonstrate the
effectiveness of our method,
yielding error-correction rate of 80.46\%, and improved recognition
rate of 95.53\% from before-post-processing rate 71.2\% for single
best-solution selection.

\clearpage

\begin{abstract}
Most of the post-processing methods for character recognition rely on  
contextual information of character and word-fragment levels. 
However, due to linguistic characteristics of Korean, such low-level  
information alone is not sufficient 
for high-quality character-recognition applications, and we need much higher-level 
contextual information to improve the recognition results. This paper 
presents a domain independent post-processing technique that utilizes 
multi-level morphological, syntactic, and semantic information as well 
as character-level information. The proposed 
post-processing system
performs three-level processing: candidate character-set selection,
candidate {\em eojeol} (Korean word) generation through morphological
analysis, and final single
{\em eojeol}-sequence selection by linguistic evaluation. All the 
required linguistic 
information and probabilities are automatically acquired from a   
statistical corpus 
analysis. Experimental results demonstrate the effectiveness of our method,
yielding error correction rate of 80.46\%, and improved recognition
rate of 95.53\% from before-post-processing rate 71.2\% for single
best-solution selection.  
\end{abstract}

\section{Introduction}
Optical character recognition has been actively researched as convenient
means of automatic data input to computers. However, due to the 
similarities among recognized characters and noises in images, there 
have been limitations to the performance improvement of character recognition
method. Since humans can understand noise-contained images using lexical and
grammatical knowledge, character recognition systems also must utilize the 
contextual information via post-processing of recognized characters.
Post-processing can improve the overall recognition performance by 
correcting the recognition errors and selecting the
most appropriate characters among the several candidates according to the
given contexts. 

Previous post-processing methods \cite{srihari:computer} use only 
within-word contextual information
such as character transition and confusion probabilities
based on Markov assumptions to perform Viterbi-style 
searches. They also
use character similarity metrics to do some dictionary search, and some
systems use morphological analysis to find the word structures. However, all
these systems have limitations that they only use within-word contextual
information, and do not utilize between-word/phrase information. 
In contrast, we extend the contextual information to sentence level, 
and present a 
multi-level post-processing method that utilizes linguistic information 
including character\footnote{The character
in Korean character recognition actually designates a syllable in  
linguistic terminology. 
Korean character recognition is performed on syllable-based, rather than
alphabet-based as in English, because Korean writing system enforces
a two-dimensional syllable structure. When we mention a character
regarding to our system in this paper, the readers should know that we 
actually mean a syllable.}, word\footnote{The Korean word is a
group of clearly distinguishable morphemes, and is called an {\em eojeol}. 
Korean is an agglutinative language which has very complex word structure.
In this paper, we will interchangeably use the term 'word' and {\em eojeol}.}, 
syntax, and even semantic-based information for domain-independent off-line 
text recognition. The proposed post-processing system
performs three-level processing: candidate character-set selection,
candidate {\em eojeol} (Korean word) generation through morphological
analysis, and final single
{\em eojeol}-sequence selection by linguistic evaluation. All the
required linguistic
information and probabilities are automatically acquired from  
a statistical corpus analysis.

The paper is
organized as follows. Section~\ref{sec:related} surveys previous
approaches for post-processing and their limitations. In
section~\ref{sec:linguistic}, we introduce the high-level 
linguistic information employed for our post-processing method. 
Section~\ref{sec:multi} shows the architecture of the system, and 
explains the multi-level post-processing method in detail. 
Section~\ref{sec:experiment} demonstrates the effectiveness of our method by
showing several experimental results and analyses, and  
finally section~\ref{sec:con} draws some conclusions.

\section{Related researches}
\label{sec:related}
Previous post-processing methods mostly utilized character-level
contextual knowledge. According to the contextual knowledge representation,
these post-processing methods can be classified as bottom-up (data-driven),
top-down (knowledge-driven), and bottom-up/top-down hybrid approaches.
In the bottom-up methods such as Viterbi algorithm
or modified Viterbi algorithm \cite{shinghal:experiment,srihari:integrating},
the contextual knowledge is represented probabilistically using Bayesian
formalism and Markov assumptions. The algorithm searches for the most-likely 
solution character sequences given the 
recognized characters using prior and
conditional (confusion) probabilities. The Viterbi algorithm is efficient,
but can generate solutions which are not in the given dictionary,
which yields relatively low error-correction performance.  
The top-down methods directly search the dictionary 
to find the most similar character sequences given the recognized 
sequences \cite{toussaint:use}. The dictionary search method usually 
guarantees good error-correction performance, but also suffers from high costs.  
The dictionary can be approximated using the binary n-gram (BNA) technique
\cite{ullman:binary}. The BNA dictionary can be used to find if the 
recognized word contains errors, and also the position of the errors. 
The BNA technique can also correct the errors, and is more efficient than
the direct dictionary search. However, BNA performance is degraded when the
word length is short, and the technique generates too many correction candidates. To
overcome the limitations of both top-down and bottom-up methods, some hybrid
methods are also suggested \cite{shinghal:bottom}. These methods basically
try to exploit both the efficiency of Viterbi search and the performance 
of dictionary look-ups. All these previous researches for English language try to 
find the best solution-character sequences using the character-level information, and
rarely try to utilize the more high-level linguistic constraints. 
However, Korean
is an agglutinative language which has very complex word structure, and has
two-dimensional syllable-based writing systems. So all these
character-based error-correction schemes are too narrow scoped, and cannot
give a good performance since Korean recognition should be syllable-based,
rather than character-based. 

Considering these characteristics of Korean, some researches on Korean character
recognition have used morphological analysis and various kinds of linguistic
assessments. 
Lee et. al. \cite{lee:korean} used several dictionaries and morphological analysis techniques 
to correct Korean spelling errors. Their dictionaries consist of morpheme 
dictionaries and inverse dictionaries of functional words (noun-endings and
verb-endings). 
Later, they extended their methods to incorporate various linguistic heuristics 
to develop
error-type decision functions, and obtained 77.5\% of error-correction rate
\cite{sim:morph}. However, they didn't use any statistical information and solely
depended on the symbolic heuristics, therefore yielding error-prone and fragile systems. 
Hong et. al. \cite{hong:study} used morphological analysis and binary n-gram (BNA) techniques for detecting and
correcting errors. Their method showed great efficiency in correcting mis-recognized
and un-recognized characters, but the BNA techniques are inherently weak in 
short-word error correction. Moreover, they couldn't correct 
the multiple errors occurring 
simultaneously in two or more morphemes. Lee et. al. \cite{lee:error} argue that
post-processing results should be fed-back to the feature extraction and recognition
stage. By applying syntactic word structures and character-level probabilities back
to the previous stages, they could increase their recognition rate 11\% from 86\% to 97\%.   
But the feedback can increase the system complexity and therefore tends to be more
time consuming. Due to the similar linguistic structures, morphological analysis and
linguistic evaluation have also been used in Japanese character recognition
post-processing. In \cite{sambe:correction}, they used morphological analysis to
produce all the possible candidate strings and applied evaluation functions based-on
Japanese word- or phrase-level heuristics to calculate the phrase plausibilities. 
By using the evaluation functions, they could increase their recognition rate 6.8\%
in average. Some systems used detailed domain knowledge to the error-correction and 
showed a great success. For example, Lee and Kim \cite{lee:efficient} used special dictionaries
and algorithms designed for each of province names, address numbers, building names, 
and people names in postal addresses, and obtained very good performance of error
correction. Similarly, \cite{niwa:post} also utilized domain knowledge as well as
linguistic knowledge to evaluate the plausibility of bunsetsu (Japanese word) 
candidates, and could improve the recognition rate for even very un-reliable 
recognition devices. However, these systems are domain-dependent and cannot be
compared with the general purpose post-processing systems. 

Contrary to English systems which mostly use character and word-fragment level information, 
our post-processing scheme focuses on beyond morpheme and 
between {\em eojeol} linguistic
constraints for more broad and efficient error-correction for Korean.
Unlike the previous Korean systems, our scheme utilizes both statistical and
symbolic information for efficient error-correction, and employs multi-level
feed-forward architecture incorporating all the character-level, morphological,
syntactic and semantic co-occurrence knowledge. Each of the knowledge is used in 
domain-independent way, so our scheme can be well applied to general
texts regardless of their domains. 

\section{High-level linguistic information for post-processing}
\label{sec:linguistic}
Broadly speaking, the linguistic information used in post-processing can be
any kind of statistical or structural linguistic constraints from character
level to semantic level, or even to pragmatic level. The followings are 
summary of linguistic
constraints that can be utilized in character recognition post-processing:
\begin{itemize}
\item character/word-fragment level: character confusion probabilities, 
character transition probabilities, and character-based n-grams
\item morpheme/word level: word structure information (morphotactics) 
and lexical frequencies
\item syntax level: structural or statistical relations between words/phrases
including part-of-speech tags
\item semantic level: semantic selectional restrictions, and word co-occurrence
relations
\end{itemize}
Our post-processing extends the linguistic information up to the semantic
level for practical post-processing performance, especially for off-line
printed character recognition for massive texts.
This section explains the high-level linguistic information (syntax and
semantics level) for the post-processing. These linguistic constraints  
provide the basis for linguistic evaluation during the multi-level 
post-processing. 

\subsection{Part-of-speech tags and tagging}
A single word can usually have multiple part-of-speech's (POS's)  
according to the given 
contexts, and when it is the case, we say that the word exhibits a 
POS ambiguity. 
POS tagging is a disambiguation process that assigns the most 
appropriate POS tag sequence
to a given sentence (word sequence) by utilizing the contextual information.
When the character recognition results give several
possible morphological analyses, the POS tagging can provide syntax-level 
constraints in order to delete erroneous recognition results. 
In this paper, we employ 
the tri-gram tagging model based on HMM (hidden-markov model) 
process \cite{church:stochastic}. Constructing an appropriate tagset is essential for any
tagging application, and usually the tagset must be in the proper 
granularity. Extremely refined tagset promises the best application performance, but the
tagset tends to be impractical in size. We use a total of 20 tags for morphemes as
shown in table~\ref{tb:tagset}. 
Since Korean word (called {\em eojeol}) usually consists of 
two or more morphemes, an {\em eojeol} tag becomes a concatenation of
the constituent morpheme tags. 
    
\tabhere{tb:tagset}

The tagging unit can be a morpheme or an {\em eojeol} in Korean. However, since the
morphological analysis already provides the constraints between morphemes, we
adopt an {\em eojeol} as our tagging unit to obtain the necessary syntactic constraints
for character recognition post-processing. The tri-gram tagging model 
computes the best tag
sequence $t_{1,n}$ that satisfies the equation~(\ref{eq:tag}) in a given 
sentence. The sentence is composed of morphologically analyzed {\em eojeol} 
sequence $e_{1,n}$. 
\begin{equation}
\label{eq:tag}
T(e_{1,n}) = argmax_{t_{1,n}} p(t_{1,n} \mid e_{1,n})
\end{equation}
Using Bayesian reformulation to drop the constant {\em eojeol} sequence 
probability, and applying two Markov assumptions to the resulting joint
probability that 1) the
current {\em eojeol} only depends on the current tag, and 2) the current tag 
depends on the previous two tags, equation~(\ref{eq:tag}) can be transformed 
into equation~(\ref{eq:tag2})\footnote{The boundary condition should be
considered when $i=1$ and $i=2$ in this equation.}. 
\begin{equation}
\label{eq:tag2}
T(e_{1,n}) = argmax_{t_{1,n}} \prod_{i=1}^{n} p(e_{i} \mid t_{i}) 
p(t_{i} \mid t_{i-2,i-1})
\end{equation}
In equation~(\ref{eq:tag2}), $p(e_{i} \mid t_{i})$ and
$p(t_{i} \mid t_{i-2,i-1})$ are called lexical probability and contextual
probability respectively, and these probabilities can be estimated by the
frequency counts from a corpus as follows:
\begin{equation}
\label{eq:tag3}
p(e_{i} \mid t_{i}) = \frac{freq(e_{i},t_{i})}{freq(t_{i})}
\end{equation}
\begin{equation}
\label{eq:tag4}
p(t_{i} \mid t_{i-2,i-1}) = \frac{freq(t_{i-2,i})}{freq(t_{i-2,i-1})}
\end{equation}
Using these two frequency count estimations, Viterbi algorithm 
is applied to search the optimal tag sequence satisfying
equation~(\ref{eq:tag2}) efficiently in polynomial time. 

\subsection{Co-occurrence patterns}
\label{sec:mut}
A word which exhibits specific meaning tends to occur in a certain context with 
other specific words, and the phenomenon is called
co-occurrence relations. For example, in Korean, the word {\em ip}\footnote{Yale
romanization is used for Korean alphabets through out in this paper.} (mouth)
usually occurs with the word {\em ta-mwul-ta} (shut). Even if the word {\em
ta-mwul-ta} has the meaning of "shut", it cannot occur with the word
{\em mwun} (door) in Korean. The co-occurring word pairs develop
co-occurrence patterns, which can give semantic constraints for the
recognized words in a sentence. There have been many researches for
automatically extracting co-occurrence patterns from a corpus in several
application areas \cite{yamashina:collocation,smadja:automatic}. 
We want to use the co-occurrence patterns as semantic constraints to
disambiguate the several candidate {\em eojeols} in the post-processing. 

There are two types of co-occurrence relations used in our post-processing
system.
The first relation is between predicates\footnote{In Korean, verb, adjective
and verbalized nouns (noun + predicate-particle) are used as predicates.} 
and their nominals, which can be used as the predicate-argument 
selectional restrictions.  We do not use any structural information that 
requires any form of parsing process  
to extract the co-occurrence relations. The post-processing mostly needs a 
lexical disambiguation, rather than a structural one, so the parsing 
overhead cannot be traded
off in the efficient character recognition post-processing. Moreover,
current parsing technology is not robust enough to handle 
unrestricted texts. Instead, we just
simply extract the {\em eojeols} and the part-of-speech's to represent the 
co-occurrence relations. The second co-occurrence patterns occur 
between two mutually
associated nominals. For example, the word {\em un-hayng} (bank) usually 
occurs with the word {\em ton} (money). In this case, we usually take into account  
the words which are associated with only limited number of
other words. If a word tends to occur with so many other words, then the word    
is too general to be associated with any specific word, and the
co-occurrence patterns become meaningless in this case. The degree of
word generality can be calculated using the following generalization
factor: 
\begin{equation}
generalization\ factor = \frac{the\ number\ of\ co-occurring\ words} 
{frequency\ of\ the\ word\ itself} 
\end{equation}
We only consider the words with small generalization factor to extract the 
meaningful co-occurrence patterns. 

The co-occurrence relations can be quantified by calculating mutual
information among the co-occurring words. The mutual information is an
information-theoretic measure of the word association, and can be calculated
based on a corpus. The mutual information I(x,y) between two words x and y is 
defined as in equation~(\ref{eq:mi}) \cite{sambe:correction}.
\begin{equation}
\label{eq:mi}
I(x,y) = log_{2} \frac{p(x,y)}{p(x)p(y)} \approx log_{2} \frac{Nf(x,y)}{f(x)f(y)}
\end{equation}
In equation~(\ref{eq:mi}), p(x) and p(y) designate word occurring
probabilities, and p(x,y) is a joint occurring probability of the two 
words x and y.
The probabilities can be approximated using the word occurring frequencies f(x)
and f(y), and the joint occurring frequencies f(x,y) within a sentence, all
of which can be acquired
from a corpus of size N. The calculated I(x,y) has bigger values when the
two words are strongly associated. In that case, the co-occurrence patterns
exhibit more strong semantic constraints for the post-processing.

\section{Multi-level post-processing}
\label{sec:multi}
\begin{figure}
\centerline{\psfig{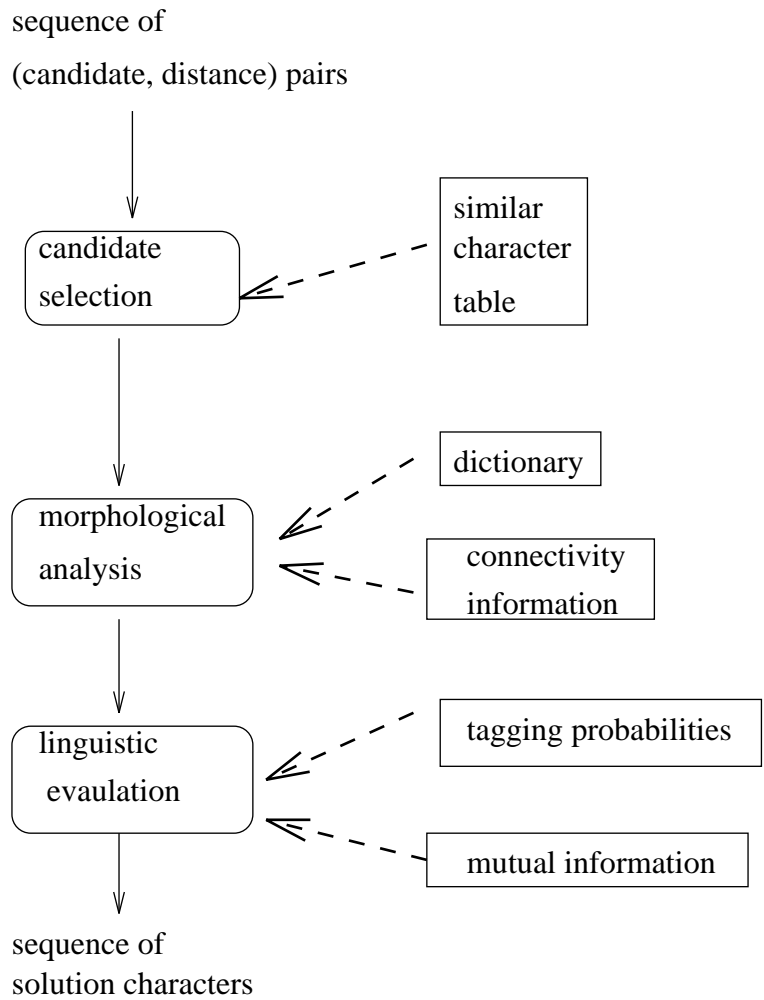}}
\caption{The architecture of multi-level post-processing}
\label{fg:arch}
\end{figure}

Basic purpose of the post-processing is a disambiguation of multiple
recognition results. In our system, input to the post-processing is a 
recognition result which consists of (candidate, distance)
pairs for each character (Korean syllable). The distance is a normalized 
recognition score between an input pattern
and its candidate pattern, and becomes smaller
when the recognition accuracy gets higher. Among the recognized candidates,
the post-processing selects
the best candidate character in a given context by applying multi-level
constraints in order to delete the inappropriate recognition results.
Applying multi-level constraints is especially necessary for Korean character
recognition because Korean recognition is syllable-based, not single character-based
such as in English recognition. If we only apply character-level probabilistic 
information, we
cannot cope with the complex word structures. The Korean dictionary needs a morpheme
as a header so the probabilistic dictionary look-up for closest word match is not
efficient because it requires word-based dictionary search. We adopt a multiple
filtering scheme that selects the final solutions step-by-step among all the
possible candidates.   
Figure~\ref{fg:arch} shows our multi-level post-processing architecture.

\fighere{fg:arch}

The candidate selection uses character-level information to restrict 
the number of candidates for simplicity of later
processing, and also to supplement the candidate sets by adding similar 
characters for
error correction. The morphological analysis uses word-fragment level constraints to
filter out erroneous recognition results by checking if the recognized 
character sequences can form grammatically correct {\em eojeols}. The
linguistic evaluation uses syntax and semantic-level statistical information
to further filter out the erroneous results. 

\subsection{Candidate selection}
\subsubsection{Candidate restriction}
\label{sec:canres}
The recognition device produces many candidates for each character, so the
character combinations can exponentially increase in a word. 
The excessive
number of candidates increases the post-processing time and decreases the
overall recognition rates due to excessive false alarms in the dictionary
look-up. The candidate restriction is performed based on the recognition
score of the best scored  
candidate (called the first candidate) for each character. If the score 
of the first candidate is very high, then many candidates can be
curtailed safely because the character is well-recognized in this case. 
To formulate the candidate restriction process, suppose $S_{0}$ is a set of 
(candidate, distance) pairs for a character, sorted by increasing 
order of the distance.  
\begin{equation}
S_{0} = \{(c_{1},d_{1}), (c_{2},d_{2}), \ldots (c_{n},d_{n})\} 
\end{equation}
where $c_{i}$ and $d_{i}$ are the i-th candidate and distance. The result of
the candidate restriction can be represented in $S_{1}$:
\begin{equation}
S_{1} = \{(c_{i},d_{i}) \mid (c_{i},d_{i}) \in S_{0}, d_{i}-d_{1}<\theta_{1},
\frac{d_{i}-d_{1}}{d_{1}}<\theta_{2} \}
\end{equation}
where $\theta_{1}$ and $\theta_{2}$ are thresholds of the restriction which
should be determined to reflect the characteristics of the recognition device. 

\subsubsection{Candidate supplement}
The candidate supplement is required for very similar characters which 
are almost impossible to be distinguished by using only the pattern themselves. 
Especially, Korean has a lot
of similar characters that result in frequent recognition errors
\cite{do:similar}. For each mis-recognizable character, candidate supplement 
recovers recognition errors by inserting its similar characters into its 
candidate set. 
We use the similar-character table for Korean in which mutually  
mis-recognizable characters are collected in pairs. The
similarity between characters was determined by the experiments
\cite{do:similar}. The candidate supplement process can be formulated as 
follows: 
\begin{equation}
S_{2} = S_{1} \cup \{(c,d_{i}) \mid (c_{i},d_{i}) \in S_{1},\ (c_{i},c) \in
similar-character\ table\}
\end{equation}
To prohibit the excessive increases in candidate numbers, currently we only supplement
the first candidates that have the minimum distance. 
Figure~\ref{fg:raw} shows the output of the recognition device that produces
10 candidates for each recognized character.

\fighere{fg:raw}

After performing the candidate restriction and supplement, the candidate set
is like in figure~\ref{fg:can}.

\fighere{fg:can}

\subsection{Morphological analysis}
The morphological analysis segments an {\em eojeol} into a sequence of morphemes,
and recognizes the constituent morphemes' root forms from phonological changes. 
Usually many morpheme combinations are possible in a single {\em eojeol}, so
we must have the knowledge of morphotactics to extract only grammatically
correct  
morpheme combinations. The morphological analysis also must handle the
phonological changes such as irregular conjugation, hiatus, contraction, and
so on. The morphological analysis can play important roles in
character recognition post-processing since it can filter out erroneous
recognition results by checking if the sequence of recognized characters can
form a grammatically correct combination of morphemes (that is, 
an {\em eojeol}). 
We developed an Korean morphological analyzer based on a tabular parsing
method \cite{aho:theory}. The algorithm utilizes two linguistic
resources: a trie-structured morpheme dictionary and a connectivity-information
table. The dictionary encodes the hierarchically organized and 
morpho-syntactically refined part-of-speech (POS) symbols\footnote{From the basic
part-of-speech, we developed very fine grained categorization of every
Korean morpheme (about 400 categories). These 400 fine grained category symbols 
are used for the Korean morphotactics modeling. Note that the tag-set used in POS
tagging is a subset of these 400 category symbols.} for each morpheme entry,
and the 
connectivity-information table encodes all the possible combinations between
these POS symbols. The morphological analysis should be
performed on every sequence of characters that can be formed by
permutations of each recognition candidate. However, since the number of
possible sequences grows exponentially, we organize the dictionary in the trie
structure \cite{aho:data} to utilize the trie's prefix-closed 
property, that is, if a string is in a trie, then
all the prefixes of the string must also be in the trie. Since our 
morphological analysis is performed by scanning from right to left, 
the trie actually contains reverse strings of the morphemes.

The morphological analysis based on the tabular parsing consists of two 
important processes: dictionary search and connectivity checking
(see figure~\ref{fg:morph}).  
The dictionary search extracts all the possible 
morphemes in an {\em eojeol}, and the connectivity checking deletes out all
the grammatically incorrect morpheme combinations. 

\fighere{fg:morph}

\begin{figure}
\centerline{\psfig{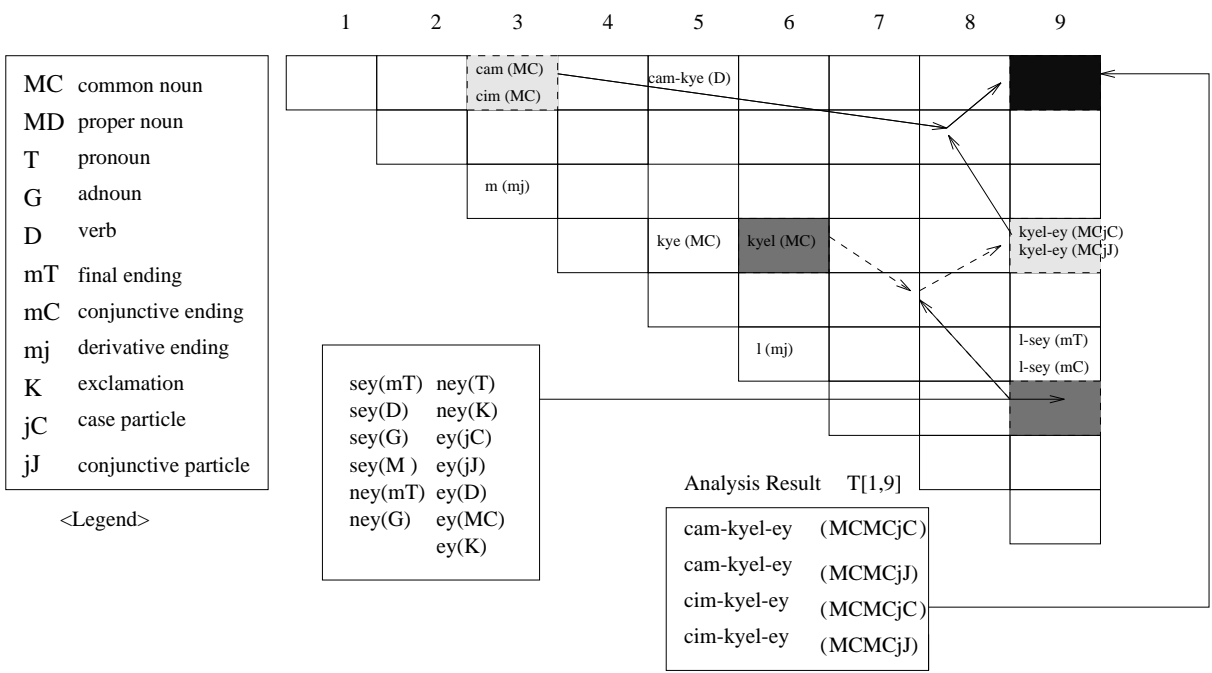}}
\caption{Morphological analysis results for the first eojeol {\em camkyel-ey}
in the recognized sentence. Among the total 8 {\em eojeol} candidates,
only the morphologically
correct sequences (with each 2 different POS tags) remain in
the final position T[1,9].}
\label{fg:table}
\end{figure}

The dictionary search position is controlled using the triangular-table
where T[i,j] holds the morphological analysis results between i-th 
and j-th character in an {\em eojeol}. The T[i,j] can
be formed either by a single morpheme or by a combination of morphemes in 
the T[i,k] and T[k+1,j], where k is between i and j-1. So the algorithm is in
principle a dynamic programming technique. Figure~\ref{fg:morph}
shows the description of the algorithm, and figure~\ref{fg:table} shows a
example morphological analysis result in the triangular-table. Since all the partial
results (intermediate combinations of the morphemes) are in the position of 
the last column, the actual time complexity   
is $O(n^2)$ at worst case when n is the number of characters in the 
input {\em eojeol}. However, since the trie property can access all the
prefixes of the found string at once, the actual dictionary access 
time is $O(n)$. 

\fighere{fg:table}

\subsection{Linguistic evaluation}
The morphological analysis usually selects several morphologically-correct {\em
eojeols} in a sentence, but not all of them are correct in
the given syntactic and semantic contexts. As the final level of
post-processing, we score each {\em eojeol} according to the high-level
linguistic constraints, and select a single correct {\em eojeol} depending on the
scores. The high-level linguistic constraints used are syntactic-level 
tagging scores and semantic-level co-occurrence scores. Since the tagging
and the co-occurrence relations are already explained in 
section~\ref{sec:linguistic}, this
section only illustrates how the two linguistic constraints are actually
applied to the post-processing.   

\begin{figure}
\centerline{\psfig{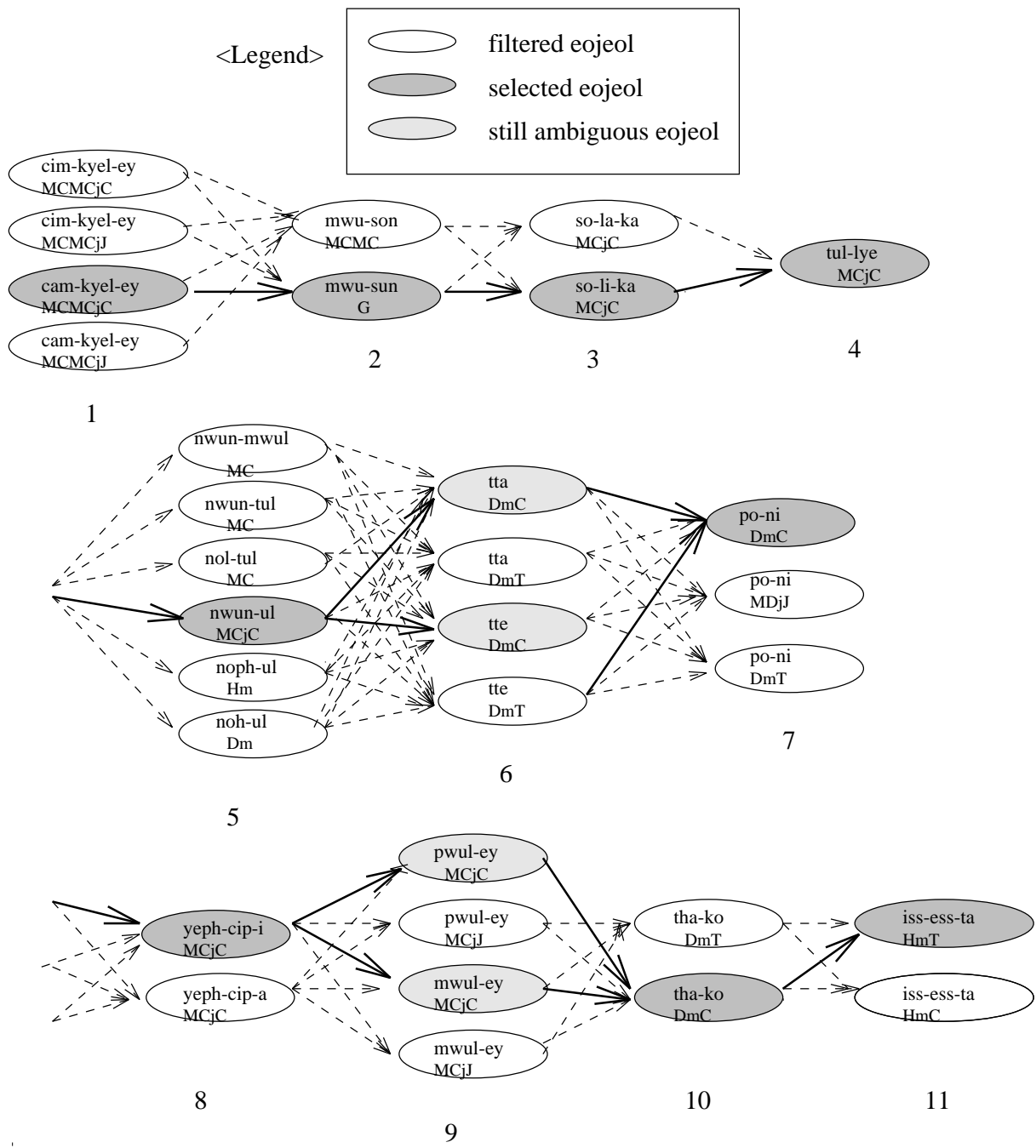}}
\caption{The ambiguity resolution using the tri-gram tagging after
morphological analysis. The numbers represent the {\em eojeol} positions in
the sentence.}
\label{fg:tagres}
\end{figure}

Figure~\ref{fg:tagres} shows how tri-gram tagging filters out  
implausible candidate {\em eojeol} sequences. 

\fighere{fg:tagres}

Since the tagging only relies on the syntactic-level constraints that are
manifested by the {\em eojeol} lexical probabilities and 
transition probabilities, there still remain semantic ambiguities even in
the best tagging paths as shown in figure~\ref{fg:tagres} 
(represented as the solid arrows). 
For the safe pruning, we select the n-best
tagging paths and deliver the multiple results to the semantic co-occurrence
checking process.  
The co-occurrence patterns can help produce further 
semantically-disambiguated {\em eojeols} after the tagging process. 
This process works especially well when the nominals or
predicates are in the ambiguous {\em eojeols}. The mutual information (see
section~\ref{sec:mut}) for the nominals (or predicates) between in the ambiguous
{\em eojeols} and in the previously disambiguated {\em eojeols} is calculated, 
and the best scored {\em eojeols} can be selected. For example, in
figure~\ref{fg:tagres}, the mutual information gives the final disambiguated
results {\em tte} and {\em pwul-ey} at the 6th and 9th {\em eojeol}
positions among the still ambiguous results (designated by the light dark circles).     
Even after the high-level linguistic evaluation, there is a chance that the
ambiguity still remains. In that case, we select the final {\em eojeol} that
has the smallest distance sum according to the candidate order from the
recognition device.

\section{Experiments}
\label{sec:experiment}
\subsection{Experiment set-up}
\begin{figure}
\centerline{\psfig{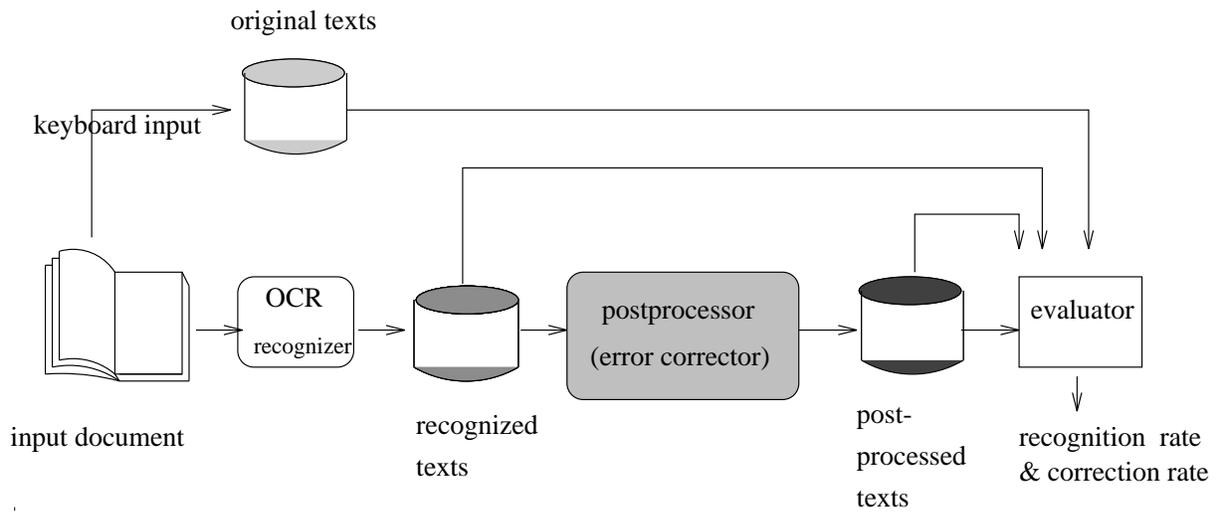}}
\caption{The experiment set-up for the multi-level post-processing}
\label{fg:exp}
\end{figure}

The experiment set-up for the multi-level post-processing is shown in
figure~\ref{fg:exp}. The original texts, recognized texts, and
the post-processed texts are compared one another to obtain the 
recognition rate and the correction rate.  

\fighere{fg:exp}

For the post-processing experiments, the following resources have been  
prepared:
\begin{itemize}
\item dictionary: a trie-structured dictionary with about 30,000
morphemes, and a connectivity information table.  
\item similar character table: character (syllable) similarity is calculated 
from each phoneme (consonant and vowel) similarity and the recognition device
confusion probabilities. We constructed about 100 entries of similar character table for Korean.
\item tagged corpus: lexical and transition probabilities for tagging,
and mutual information for co-occurrence patterns are acquired from a tagged
corpus. We built a tagged corpus using about 3,000 sentences (23,000 {\em
eojeols}) from elementary-school textbooks and raw sentences supplied from
ETRI\footnote{Electronics and telecommunications research institute in
Korea}. From this tagged corpus, we extracted about 140 uni-grams, 1,300
bi-grams, and 5,000 tri-grams for {\em eojeol} tags. 
\item test data: the 1,722 {\em eojeol} test data are selected from the
elementary-school textbooks, and divided into 3 sets A, B, C according to the 
OCR recognition rate (68.4 \%, 69.5 \%, 75.6 \% respectively). 
\end{itemize}

\subsection{Experiment results and analyses}
\subsubsection{Performance measures}
We use correction rate and recognition rate (after post-processing) for 
our performance measures.  The correction rate is defined as follows: 
\begin{equation}
correction\ rate = \frac{(successfully\ corrected\ characters) - 
(mis-corrected\ characters)}
{total\ erroneous\ first\ candidates} \times 100
\end{equation}
where "mis-corrected" means that the correctly recognized characters become
incorrect due to the post-processing. On the other hand,   
the recognition rate is defined as follows: 
\begin{equation}
recognition\ rate = \frac{correctly\ recognized\ characters}
{total\ first\ candidates} \times 100
\end{equation}
Figure~\ref{fg:expc} and figure~\ref{fg:expe} shows the correction rate and
the recognition rate (before and after post-processing) for characters and {\em
eojeols} with each document set A,
B, C and their average. The correction rate is high when the original 
recognition rate (before post-processing) is high. This means that the 
error correction performs well for the highly confident candidate sets that
have small distances. However, the overall recognition rate after
post-processing generally becomes high even for the low original recognition 
rate, so the post-processing can be practically used for the low 
recognition-rate devices. 

\fighere{fg:expc}

\fighere{fg:expe}

\subsubsection{Candidate selection effects}  
\begin{figure}
\centerline{\psfig{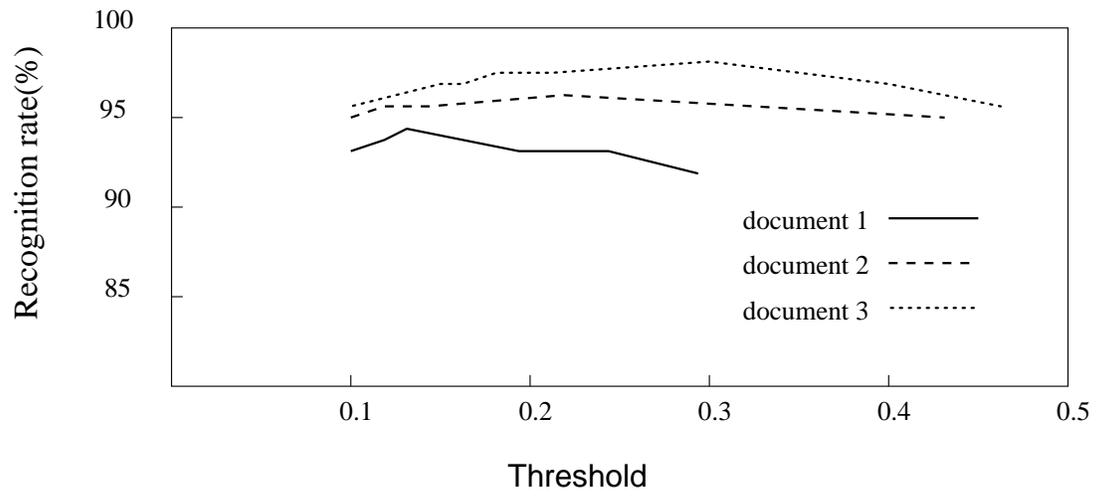}}
\caption{Candidate restriction threshold and recognition rate}
\label{fg:canres}
\end{figure}

The post-processing is performed on sentences, so the processing time
depends on the number of candidate {\em eojeols} generated by the morphological 
analysis and the sentence length. The candidate {\em
eojeols} are composed of candidate character combinations, so the processing
time exponentially increases according to the number of candidate
characters. Too many candidate characters also degrade the recognition rate
since the {\em eojeols} made of low order candidates might get high scores 
in the linguistic evaluation. However, too few candidates might result in no
solution in the candidate character set. Figure~\ref{fg:canres} shows the
effect of candidate restriction by showing the recognition rate according to
the threshold $\theta_1$ (see section~\ref{sec:canres}).

\fighere{fg:canres}

\begin{figure}
\centerline{\psfig{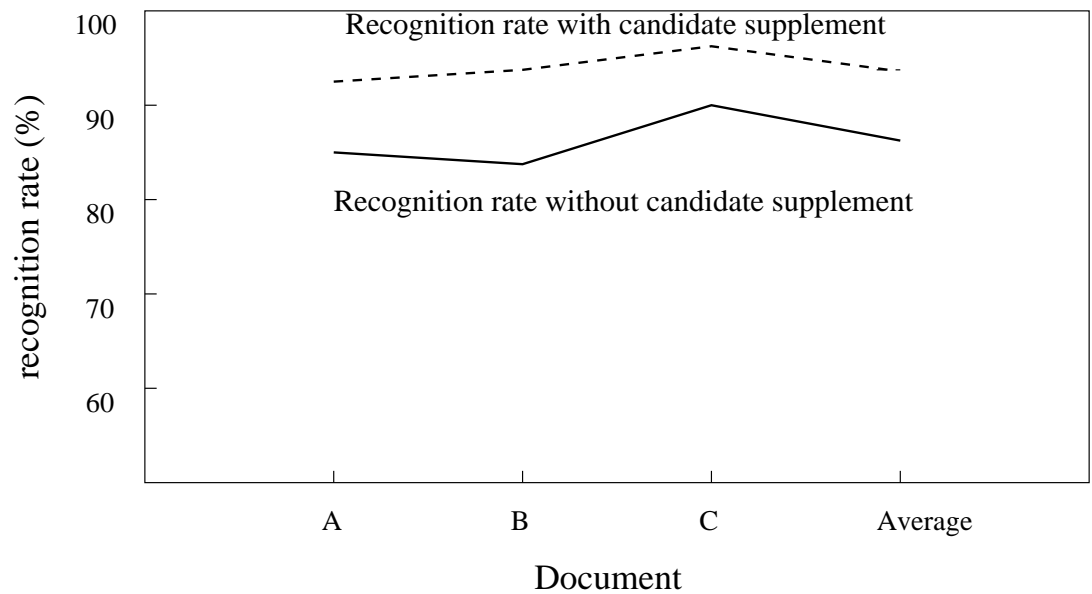}}
\caption{The candidate supplement effects using similar character sets}
\label{fg:cansup}
\end{figure}

As shown in the figure, the number of candidates that yields the best recognition
rate depends on the document sets, hence on the recognition devices. We have to
choose the best threshold values according to recognition devices
through experiments.
The post-processing assumes that there is at least one correct solution in the
candidate set. However, in reality, the Korean character set (2350
different characters) contains so many similar characters that there might 
not be  any solution character in
the candidate character set. Therefore, we supplemented the first candidate to
include all the similar characters according to the device confusion
probabilities and the original character similarity. Figure~\ref{fg:cansup} 
shows the candidate supplement effects.

\fighere{fg:cansup}

\subsubsection{Ambiguity resolution performance}
\begin{figure}
\centerline{\psfig{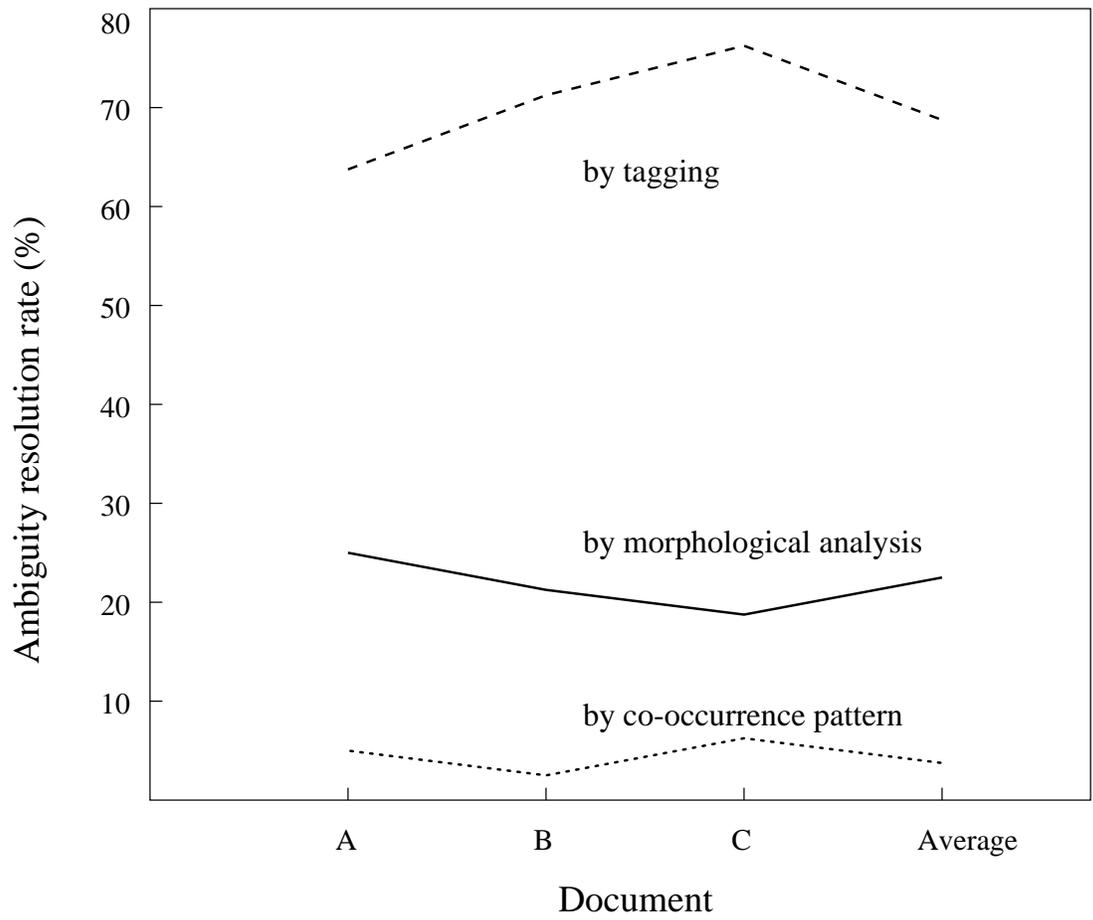}}
\caption{Disambiguation performance of each linguistic constraint}
\label{fg:dism}
\end{figure}

The post-processing process can be interpreted as a disambiguation process
that selects a single solution character among several candidate characters.
We apply multi-level linguistic constraints for the disambiguation of
characters in an {\em eojeol} structure.
Figure~\ref{fg:dism} shows a disambiguation performance of each linguistic
constraint application: the morphological analysis, the tagging, and the co-occurrence 
patterns. 

\fighere{fg:dism}

The ambiguity resolution rate for each specific linguistic processing is 
defined as follows:
\begin{equation}
ambiguity\ resolution\ rate = 
\frac{recognition\ rate\ increase\ after\ the\ specific\ linguistic\ processing}
{total\ recognition\ rate\ increase}
\end{equation}
Even after applying all the linguistic constraints, about 4\% test data still 
have ambiguities. So we had to decide the final solutions based on the candidate 
order from the recognition device.  

\subsection{Discussions}
The linguistic information used is statistical, rather than structural, so
it can be automatically extracted from a corpus and is robust in its nature.
However, the statistical information inherently depends on the corpus, so
the words which are not in the training corpus result in zero frequencies
in the post-processing. So some form of smoothing is always necessary to deal 
with
this sparse data problem. For the tri-gram tagging, we used the uni-gram and
bi-gram together for the smoothing\cite{charniak:stat}.
\begin{equation}
p(t_{i} \mid t_{i-2,i-1}) = \lambda_{1}p(t_{i}) + \lambda_{2}
p(t_{i} \mid t_{i-1}) + \lambda_{3}p(t_{i} \mid t_{i-2,i-1})
\end{equation}
where $\lambda_{1}+\lambda_{2}+\lambda_{3}=1$. The sparse data problem also
generates the zero co-occurrence frequences in mutual information
calculation, and results in -$\infty$ in the value. Basically, this problem 
can be handled with the semantic category-based mutual information using a
well-developed thesaurus. However, well-developed Korean thesaurus is not
available at the moment, so we had to develop another smoothing technique.
In order to cover the words that do not co-occur in the training data, 
we employ the
single word frequencies together with the mutual co-occurrence frequencies
such as: 
\begin{equation}
I_{new}(x,y) = \lambda_{1}(f(x)+f(y)) + \lambda_{2}I(x,y)
\end{equation}
Usually the co-occurrence pattern size is enormous when we consider all the
(predicate, nominal) and (nominal, nominal) pairs in the dictionary.
However, our system only extracts the co-occurrence patterns for the words
that occur in the real corpus. Moreover, we only consider the words that
have more than a certain amount of actual co-occurrence frequencies, and
that have restricted number of accompanying words using the generalization
factor (see section~\ref{sec:mut}). 
Our experiments use 22,000 {\em eojeol} corpus that includes about 2,000 predicates
and 900 common nominals. According to our scheme, only about 2,800
co-occurrence patterns were actually extracted among the theoretically possible more
than billion word pairs. 

\section{Conclusions}
\label{sec:con}
This paper proposes a practical post-processing system for optical character
recognition, which utilizes
high-level linguistic information as well as character-level information. 
Our post-processing method is especially useful for the applications that
require beyond word-level contexts to improve the recognition results,
such as off-line massive text recognition. 
Unlike most of the previous post-processing schemes that utilize only character and
word-fragment level information, our post-processing is executed in 3 
stages: candidate character-set selection, candidate {\em eojeol} 
generation through 
morphological analysis, and final single eojeol-sequence selection by 
the high-level linguistic evaluation. 

The candidate selection uses the distance generated by the recognition device,
and restricts the number of candidates for later
processing, and supplements the candidate sets by adding similar characters
for error correction. For the selected candidate characters, the morphological 
analysis generates only the morphologically-correct {\em eojeol} sequences 
by checking if the recognized
character sequences can form a grammatically-correct {\em eojeol}. The
generated {\em eojeols} are now grammatically correct, but may be
inappropriate in the given contexts. The
linguistic evaluation uses syntax and semantic-level statistical information
to further filter out the contextually-inappropriate {\em eojeols} for 
final recognition error correction. The linguistic evaluation is performed
in a cascaded way using syntactic tagging constraints, semantic 
co-occurrence constraints, and finally candidate orders from the 
recognition device. 

\begin{figure}
\centerline{\psfig{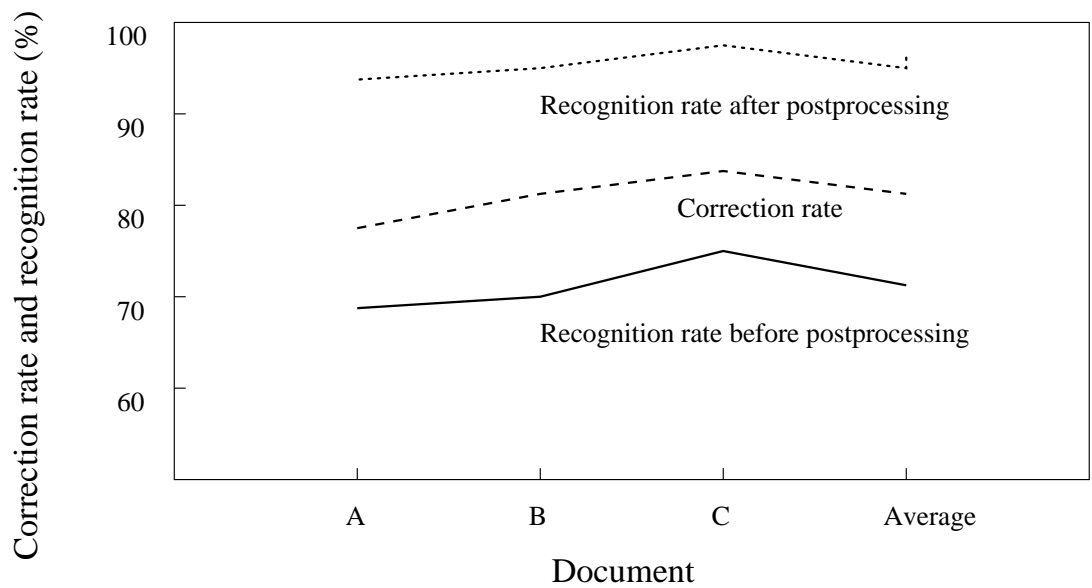}}
\caption{Recognition rate and correction rate for characters (with each test
data-set A, B, C)}
\label{fg:expc}
\end{figure}

\begin{figure}
\centerline{\psfig{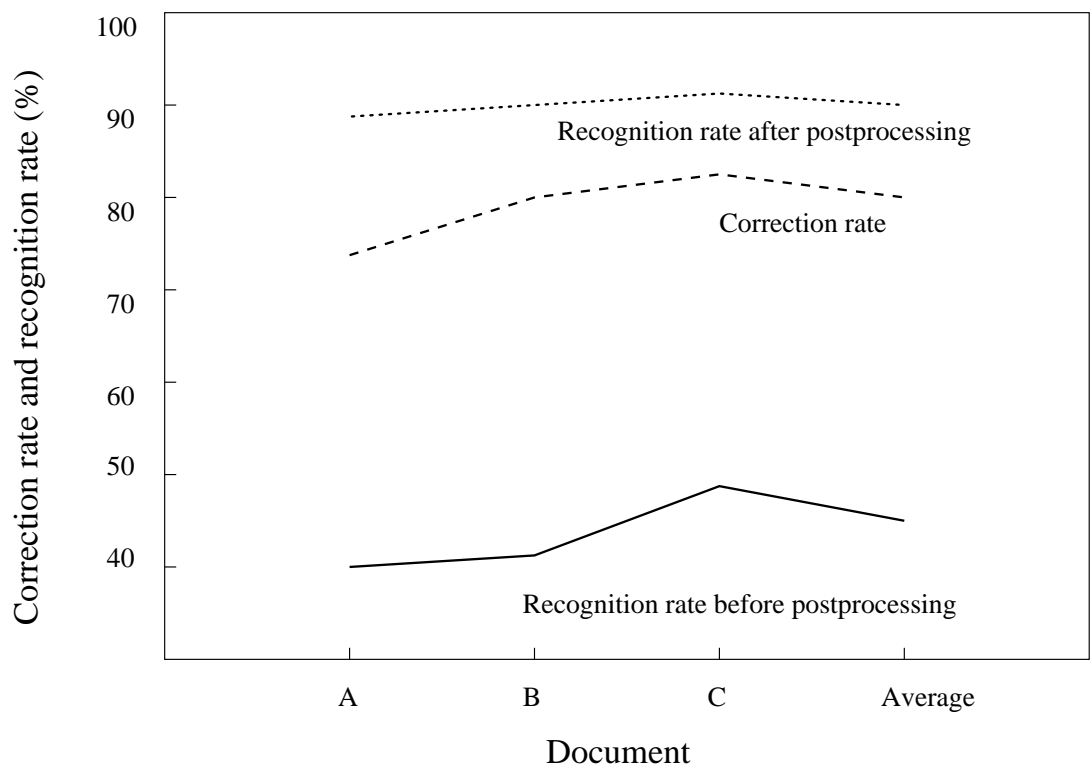}}
\caption{Recognition rate and correction rate for {\em eojeols}}
\label{fg:expe}
\end{figure}

We conducted extensive experiments to demonstrate the performance of our 
multi-level post-processing method. For the 1,722-{\em eojeol} test data 
extracted from elementary-school textbooks, we obtained 80.46\% correction rate
and 24.3\% increase of the recognition rate (from 71.2\% to 95.53\%). This performance
is much better than similar previous approaches for Korean and Japanese 
post-processing compared in section~\ref{sec:related}. Moreover, our post-processing
can be applied to any text in domain-independent way. The major
post-processing failures in our system come from the case that the selected 
candidate set
does not include the solution characters in the first place since our test-bed
recognition device is primitive and experimental one. This no solution case
propagates to the next stages of the post-processing, resulting in the morphological
analysis failures or incorrect {\em eojeol} selection which again gives
rise to the tagging and co-occurrence checking failures. The better
recognition devices should yield much better post-processing results as demonstrated
in figure~\ref{fg:expc} and figure~\ref{fg:expe}.  
The post-processing failures are also due to the limited corpus size which 
gives incomplete statistical linguistic constraints in the
tagging and co-occurrence pattern extraction. The larger-scale balanced corpus
should be provided for more practical post-processing.

\bibliographystyle{plain}

\clearpage
\begin{table*}
\begin{center}
\begin{tabular}{|c|c|c|c|} \hline
tag & description & tag & description \\ \hline
MP & proper noun & SC & ordinal numeral \\
MD & bound noun & SO & cardinal numeral \\
MC & common noun & e & prefinal ending \\
D  & verb & y  & predicate particle \\
H & adjective & mC & conjunctive ending \\
G & adnoun  & mT & final ending \\
B & adverb & mJ & derivative ending  \\
jJ & conjunctive particle  & T  & pronoun \\
jS & auxiliary particle & + & prefix \\
jC  & case particle & -  & suffix \\ \hline
\end{tabular}
\end{center}
\caption{Morpheme tagset for Korean part-of-speech tagging.}
\label{tb:tagset}
\end{table*}

\clearpage

\clearpage
\begin{figure}
\begin{verbatim}
[cim 104][cam 137][cal 185][kiph 197][chim 205][cap 210][cep 215][kil 227]...
[kyel 31][kel 75][kyeth 92][kil 120][kal 121][cil 135][cel 137][kath 200]...
[ney 181][ey 186][sey 232][ay 245][nay 289][may 346][say 359][yey 296]...
[mwu 114][twu 173][phwu 187][pwu 239][tok 277][mok 280][lwu 285][mo 307]...
[son 122][sun 168][chon 327][con 341][un 363][ton 436][lon 475][nwun 475]...
[so 120][o 445][su 453][u 520][ssu 578][no 692][swu 745][hwu 782]...
[li 163][la 172][le 232][toi 241][kwui 281][hi 286][hoi 299][mek 303]...
[ka 34][ki 95][ke 302][khi 302][ca 320][ci 352][kye 489][kki 491]...
[tul 87][tut 105][thul 160][mwul 197][tum 218][tol 219][lul 254][nul 314]...
[lye 106][le 162][the 200][mek 203][li 254][hye 255][chye 270][phe 280]...
[noph 202][nwun 266][noh 287][nol 297][lyo 304][pon 318][tal 362][mok 370]...
[tol 187][ul 210][mwul 225][sol 241][nol 267][phwul 271][dwul 275][swul 283]...
[tte 211][tta 320][ppe 488][mye 516][payk 534][tey 551][ye 558][me 577]...
[po 84][mo 282][o 284][pwu 315][u 333][yo 385][mu 395][pok 417]...
[ni 66][na 145][si 334][ne 349][sa 364][i 387][nye 409][a 455]...
[yeph 124][iph 182][ilh 197][anh 282][el 283][aph 288][teph 292][et 296]...
[ip 248][cip 251][ip 286][cap 293][yeng 294][cam 299][cing 302][ching 309]...
.....
.....
\end{verbatim}
\caption{Output of the recognition device for the example sentence: {\em
camkyel-ey mwusun soli-ka tul-lye nwun-ul tte po-ni yepcip-i pwul-ey tha-ko
issess-ta} (When I opened my eyes by overhearing something asleep, the
neighboring house was in flames).}
\label{fg:raw}
\end{figure}

\clearpage
\begin{figure}
\begin{verbatim}
[cim 104][cam 137]
[kyel 31]
[ney 181][ey 186][sey 232][ay 245]
[mwu 114][twu 173]
[son 122][sun 168]
[so 120]
[li 163][la 172][le 232][toi 241]
[ka 34]
[tul 87][tut 105][twul 87]
[lye 106]
[noph 202][nwun 266][noh 287][nol 297][lyo 304]
[tol 187][ul 210][mwul 225][sol 241][nol 267][phwul 271][dwul 275][swul 283]...
[tte 211][tta 320]
[po 84]
[ni 66]
[yeph 124][iph 182]
[ip 248][cip 251][ep 286][cap 293][yeng 294][cam 299][cing 302][ching 309]...
[a 191][i 207][e 267][ya 287]
[pwul 61][mwul 78]
[ney 284][ey 315][sey 382][ay 388]
[tha 138]
[ko 132]
[iss 168][ass 175][ess 201]
[ess 127][ass 161][iss 184][es 185]
[ma 227][ne 318][na 332][ta 277]
\end{verbatim}
\caption{The candidate selection result}
\label{fg:can}
\end{figure}

\clearpage
\begin{figure}
\begin{verbatim}
T[i,n] <- find_word_from_trie_dict, 1<=i<=n; /* fill last column */
for (start = n; start >= 1; start --) {
   if (!Empty (T[start,n])) {
     T[i,start-1] <- find_word_from_trie_dict, 1<=i<=start-1;

     for (left_start = start -1 ; left_start >= 1 ; left_start --) {
        /* now begin connectivity checking */
       foreach left_morph_chain in T[left_start,start-1]  {
         /* more than one chain */
        foreach right_morph_chain in T[start,n] {
         if Connectable (left_morph_chain, right_morph_chain)
          AddTo(T[left_start,n], concat(left_morph_chain,right_morph_chain));
        } /* for right morpheme */
       } /* for left morpheme */
     } /* for left_start */
   } /* if */
}  /* for start */
\end{verbatim}
\caption{Morphological analysis algorithm based on the tabular parsing
method}
\label{fg:morph}
\end{figure}

\clearpage

\clearpage

\clearpage

\clearpage

\clearpage

\clearpage

\clearpage

\clearpage

\clearpage

\begin{center}
{\bf\Large Biographical sketches for each author}
\end{center}

{\bf Geunbae Lee} received his B.S. and M.S. degrees in computer engineering
from the Seoul National University, Seoul, Korea in 1984 and 1986,
respectively and the Ph.D. degree in computer science from the University of
California at Los Angeles in 1991. Currently, he is an assistant professor
in the Computer Science Department, Pohang University of Science and
Technology, Pohang, Korea. His current research interests are in the field
of natural language processing, speech recognition, and intelligent agent
interfaces.

{\bf Jong-Hyeok Lee} received his B.S. degree in mathematics eduation from
the Seoul National University, Seoul, Korea in 1980, and his M.S. and Ph.D
degree in computer science from KAIST, Korea in 1982 and 1988, respectively.
Currently, he is an assistant professor
in the Computer Science Department, Pohang University of Science and
Technology, Pohang, Korea. His current research interests are in the field
of natural language processing, machine translation, Korean information
processing.

{\bf JinHee Yoo} received her B.S. and M.S. degrees in computer science from
the Pohang University of Science and Technology, Pohang, Korea in 1993 and
1995, respectively. Currently she works for SamSung Electronics Company. Her
current research interests are in the field of natural language processing
and character recognition.

\end{document}